\documentclass[conference]{IEEEtran}
\usepackage{cite}
\usepackage{amsmath,amssymb,amsfonts}
\usepackage{algorithmic}
\usepackage{graphicx}
\usepackage{textcomp}
\usepackage{xcolor}
\usepackage{booktabs}
\usepackage[nolist,nohyperlinks]{acronym}

\def\BibTeX{{\rm B\kern-.05em{\sc i\kern-.025em b}\kern-.08em
    T\kern-.1667em\lower.7ex\hbox{E}\kern-.125emX}}
\begin{document}

%\title{Intent based management in 5G networks}
% \title{Intent-based multi-agent reinforcement learning for service assurance in 5G networks}
% \title{Intent-based multi-agent reinforcement learning for closed-loops service assurance in cellular networks}

%\title{Intent-based multi-agent reinforcement learning for service assurance in cellular networks}

\title{Multi-agent reinforcement learning for intent-based service assurance in cellular networks}

\author{
    \IEEEauthorblockN{
        Satheesh K. Perepu\IEEEauthorrefmark{1}, 
        Jean P. Martins\IEEEauthorrefmark{2},  
        Ricardo Souza S\IEEEauthorrefmark{2} and 
        Kaushik Dey\IEEEauthorrefmark{1}
    }
    \IEEEauthorblockA{
        \IEEEauthorrefmark{1}Ericsson Research, India, 
        \IEEEauthorrefmark{2}Ericsson Research, Brazil\\
        Email: \{perepu.satheesh.kumar, jean.martins, ricardo.s.souza, deykaushik\}@ericsson.com
    }
}

%[ Jean - ORCID: 0000-0003-2112-3723] 

\begin{acronym}
    \acro{5G}{5th Generation Mobile Networks}
    \acro{AI}{Artificial Intelligence}
    \acro{API}{Application Programming Interface}
    \acro{BBU}{Baseband Unit}
    \acro{C-RAN}{Centralized Radio Access Networks}
    \acro{CV} {Conversational Video}
    \acro{DQN}{Deep Q-Network}
    \acro{DRL}{Deep Reinforcement Learning}
    \acro{GRU}{Gated Recurrent Unit}
    \acro{MBR}{Maximum Bit Rate}
    \acro{MDP}{Markov Decision Process}
    \acrodefplural{MDP}{Markov Decision Processes}
    \acro{ML}{Machine Learning}
    \acro{NN}{Neural Network}
    \acro{RL}{Reinforcement Learning}
    \acro{MARL}{Multi-agent Reinforcement Learning}
    \acro{RTT}{Round Trip Time}
    \acro{RRU}{Remote Radio Unit}
    \acro{TD}{Temporal Difference}
    \acro{UPF}{User Plane Function}
    \acrodefplural{UPF}{User Plane Functions}
    \acro{UE}{User Equipment}
    \acrodefplural{UEs}{User Equipments}
    \acro{URLLC}{Ultra Reliable Low Latency Communication}
    \acro{mIoT}{massive IoT}
    \acro{QoE}{Quality of Experience}
    \acro{UL}{Uplink}
    \acro{DL}{Downlink}
    \acro{KPI}{Key Performance Indicator}
    
\end{acronym}

\maketitle

\begin{abstract}

Recently, intent-based management has received good attention in telecom networks owing to stringent performance requirements for many of the use cases. Several approaches in the literature employ traditional closed-loop driven methods to fulfill the intents on the KPIs. However, these methods consider every closed-loop independent of each other which degrades the combined performance. Also, such existing methods are not easily scalable. Multi-agent reinforcement learning (MARL) techniques have shown significant promise in many areas in which traditional closed-loop control falls short, typically for complex coordination and conflict management among loops. In this work, we propose a method based on MARL to achieve intent-based management without the need for knowing a model of the underlying system. Moreover, when there are conflicting intents, the MARL agents can implicitly incentivize the loops to cooperate and promote trade-offs, without human interaction, by prioritizing the important KPIs. Experiments have been performed on a network emulator for optimizing KPIs of three services. Results obtained demonstrate that the proposed control approach performs quite well and is able to fulfill all existing intents when there are enough resources or prioritize the important KPIs when resources are scarce.

%Recently, intent-based management is receiving good attention in telecom networks owing to stringent performance requirements for many of the use cases. Several approaches on the literature employ traditional methods in the telecom domain to fulfill intents on the KPIs, which can be defined as a closed loop. However, these methods consider every closed-loop independent of each other which degrades the combined closed-loop performance. Also, when many closed loops are needed, these methods are not easily scalable. Multi-agent reinforcement learning (MARL) techniques have shown significant promise in many areas in which traditional closed-loop control falls short, typically for complex coordination and conflict management among loops. In this work, we propose a method based on MARL to achieve intent-based management without the requirement of the model of the underlying system. Moreover, when there are conflicting intents, the MARL agents can implicitly incentivize the loops to cooperate, without human interaction, by prioritizing the important KPIs. Experiments have been performed on a network emulator on optimizing KPIs for three services and we observe the proposed system performs well and is able to fulfill all existing intents when there are enough resources or prioritize the KPIs when there are scarce resources. 
\end{abstract}


\section{Introduction}

The future networks like 6G will primarily be driven by Intents from network operators. Such intents, each having a defined priority of fulfillment, may consist of one or more objectives. Hence each intent may be decomposed into more than one objectives by an Intent Manager. Within each network slice, there could be multiple such objectives and each in turn may need one or more parameters to be optimized. In telecom domain, each such objective measured by a KPI goal could be achieved by one closed loop which optimizes the respective parameter(s). Quite often these closed loops are interacting i.e. any change in one closed loop may affect the KPIs assured by another closed loops. Optimization for each objective in isolation or in some cases sequentially~\cite{paper:moysen_conflict_res} is possible through existing methods in literature, but the challenge manifests when the objectives conflict with each other and the model of the environment is not available. It is to be noted that the operator or higher domain functions, which are feeding such objectives, may not be aware of such conflicts and hence pre-planning on conflict avoidance may not be feasible in all scenarios. In such circumstances, the efficiency of future networks would depend on ability to autonomously manage multiple conflicting objectives and adapt~\cite{report:Ericsson} to the expectations based on priority of individual intents.

In our work, we simulated a conflict situation where an Intent Manager has to fulfill 3 intents for three different services, \ac{CV}, \ac{URLLC} and \ac{mIoT}. Each of these may involve optimizing multiple parameters in the network. For tractability, we have chosen packet priority and \ac{MBR} as the two parameters. Now considering a resource constrained environment, increase of packet priority might improve the \ac{QoE} of the \ac{CV} service but might degrade the packet loss of \ac{URLLC} service. Similarly, increasing \ac{MBR} for \ac{URLLC} may improve packet loss and reduce latency for \ac{URLLC} but may degrade the \ac{mIoT} and \ac{CV} service. Additionally the target for each of the services might change frequently and the model needs to respond towards the change without any additional training cycles. So in summary, the crux of the challenge is to optimize the realization of the intents, some of which may conflict with each other, in a resource constrained network setting.

A classical optimization technique is often not suitable as the model of the environment is not available and also the compute for optimization may not be available at run time with the targets for the goals changing frequently. Also any other model driven technique may not be scalable for the aforesaid reasons.

We have chosen a model free technique using \ac{MARL} to solve this problem. Here each of the agents are responsible for tuning a parameter related to the objective defined by the intent. A \ac{CV} may be optimized by at least two agents, one for packet priority another for MBR adjustments. The packet priority agent for \ac{CV} interacts with the packet priority agents for URLLC and mIoT and all these three agents learn to "Plan to Coordinate" to achieve an optimal global trade-off during the training phase. Additionally in our method, during execution phase, none of the agents need to observe actions or rewards (+/- towards goal) from another agent. Each agent can see only the aggregated Global reward and the current \ac{KPI} (parameter value) for its own service and hence this design prevents any communication bottleneck during execution.

Hence our method is able to address multiple intents simultaneously for circumstances where model of the environment may not be available.

\section{Background}

\subsection{Intent-based closed loops paradigm}

One trend in network management automation is the employment of intent-driven
closed loops. In this paradigm, humans move from direct actuating on
the network to a supervising role, where goals and objectives are 
conveyed through high-level intents and closed control loops act to fulfill
these intents.

Intents can be defined as \emph{formal specification of all expectations including
requirements, goals, and constraints given to a technical systems}~\cite{tmforumIntents}.
In summary, an intent defines the states a systems should reach, without explicit
information on how to reach those states. In natural language, an intent could take to the form of
"\emph{I want a conversational video service, where 80\% of the users have a \ac{QoE} of at least
4.0}", and based on that different closed loops would be instantiates to ensure intent 
fulfillment.

%Some key intent properties include:
%\begin{enumerate}
%    \item Declarative: an intent defines wanted states, but it does not define how the achieve it.
%    \item Measurable: system should be able to measure if objectives and goals are fulfllied based on 
%    observations from the underlying system.
%    \item Machine-Readable: intents must be understood by machines as well as humans without losing precision
%    or introducing ambiguity.
%\end{enumerate}

A closed loop can be defined as the management of an entity that have specific goals and that can 
be monitored and acted upon~\cite{etsi-zsm-closedloops}. According to ETSI ZSM, a closed loop consists
of four logical steps (e.g., monitoring, analysis, decision and execution). However, in the context of this work 
closed loops will be implemented by multiple goal-conditioned \ac{RL} agents. 
In this approach, intents will provide different goals for the
multiple \ac{RL} agents depending on their scope and the agents will perform all
necessary monitoring, decision and actuation tasks on the managed entities.

\subsection{Related Work}

The interaction and conflict handling among multiple intelligent agents or control-loops managing different
cellular network aspects have been investigated by several authors. Moysen et al.~\cite{moysen2018} provides a
comprehensive survey of several different elements related to \ac{ML} applied to the general area of
Self-organized Network Management, for the scope of this work we focus on conflict and coordination between multiple
control loops. As identified in~\cite{moysen2018}, most approaches focus on the proposition of a complex 
coordination mechanisms to handle conflict. These approaches range from applying problem specific 
heuristics~\cite{munoz2014}, a multi-step workflow to perform coordination while handling
scalability issues~\cite{rojas2020}, or defining the coordinator as a RL-agent~\cite{iacoboaiea2014}. One
common thread with all the aforementioned approaches is the fact that in all cases individual control loops
are independent greedy policies. That is, control loops or agents are either ML-based approaches trained 
independently or use case specific solutions and, therefore, requires more complex coordination. 

For the investigations presented on this work, we considered a multi-agent setup where agents are trained
in conjunction with each other to manage multiple services. In this setup, the conflicts are inherently
handled by the multi-agent setup and simple coordination mechanism is proposed to simplify the execution
and convergence.

%Theoretical research on the subject has also been done such as the proposal of adding Nash-equilibrium payoff to 
%Q-learning~\cite{hu2003}.

\subsection{Multi-agent Reinforcement Learning}

\ac{RL} comprises a set of techniques that enable an agent to learn to interact with an environment
by iteratively exploring and evaluating the outcomes of its actions. The overall goal in \ac{RL} is
to derive policies that map the observed situations perceived by the agent to actions that would
maximize the cumulative rewards received by such an agent~\cite{sutton1998introduction}.

%At every timestep $t=1,2,\dots,H$, one agent observes the current state of the system $s_t\in
%\mathcal{S}$ and by consulting its policy decides for an action $a\in \mathcal{A}$ that should maximize
%the cumulative rewards received from now onwards. In general, reward functions depend on states, $r: \mathcal{S}\mapsto\mathbb{R}$, while optimality is defined over trajectories $\tau=s_1,\dots,s_H$ with respect to the discounted cumulative rewards $R(\tau)=\sum_{s\in\tau} \gamma~r(s)$, where $\gamma\in[0,1]$ is a discount factor~\cite{sutton1998introduction}. According to this framework, an optimal policy is such that $\pi^* = \arg\max_\pi\mathop{\mathbb{E}}_{\tau\sim \pi}[R(\tau)]$.

%In single-agent scenarios with discrete action spaces, $Q$-learning via \ac{DQN} is a well-established algorithm~\cite{mnih2015human}. Such a technique works by employing \acp{NN} to approximate an unknown value-function $Q(s, a)$ that predicts the future cumulative rewards from any state-action pair. The goal is to derive policies $\pi(s) := \arg\max_{a\in \mathcal{A}} Q(s, a)$. Many \ac{RL} algorithms that followed \ac{DQN} success have consistently proven their effectiveness for a number of applications in diverse fields~\cite{sandberg2021learning, lee2021marl, martins2021policy,martins2021rainbow,xiong2019survey}.

There are many cases where multiple agents could be deployed in the environment but we need them to learn, based on context, either collaboration, competition or both. To learn these, we use \ac{MARL} techniques. However, learning the joint policy is not trivial, as agents usually have a local perception of the whole system (partially observable), and, in many cases, direct communication between them is absent. These and other characteristics define environments in which it is difficult for a learning agent to separate the effect of its actions from dynamics caused by other agents or other unobserved components.

The foremost technique to learn the collaboration between agents in independent Q-learning \cite{paper:iql}. However, a problem with this technique is that since each agent will take independent action, it will result in non-stationary environment. To overcome this problem, many value function decomposition methods are proposed like QMIX \cite{rashid2018qmix} etc. The idea behind this method is, the agents are trained based on combined value function $Q_{tot}$ which is obtained by combining the individual agent Q-values $Q_i$ using a neural network. In this work, we used QMIX to learn the collobaration between these agents.

% To deal with issues like these, many \ac{MARL} algorithms have been proposed in the literature~\cite{zhang2021multi}, here we focus on QMIX.

In QMIX~\cite{rashid2018qmix,rashid2020qmix}, the main goal is to guarantee that decentralized agents' decisions are consistent with those of a centralized counterpart. That is enforced during the training phase, allowing actuation to be performed in a decentralized fashion (centralized learning with decentralized execution). Each agent's policy $\pi_i$ ($i=1,\dots,n$) is derived from a $Q_i$ function. QMIX add to that setup a \textit{mixing} network $\mathbf{Q}(\mathbf{s},\mathbf{a})$, with joint states $\mathbf{s}=(s^1, \dots, s^n)$ and joint actions $\mathbf{a}=(a^1, \dots, a^n)$, from which a joint policy $\boldsymbol{\pi}(\mathbf{s})$ is derived. During training, the loss function directs the learning of $\boldsymbol{\pi}$ towards producing optimal joint actions while directing $Q_i$ value-functions towards $\mathbf{Q}$. This strategy induces individual policies $\pi(s^i)$ such that $\boldsymbol{\pi}(\mathbf{s}) = (\pi(s^1), \dots, \pi(s^n))$. That enables us to employ $\pi_i$ for decentralized execution, instead of the centralized $\boldsymbol{\pi}$.

\subsection{Intent-aware policies via Goal-conditioned RL}
In scenarios such as intent-based service assurance, we require \ac{RL} policies that can adapt to changes in the goals during the execution phase. As an example, suppose an agent is pursuing a certain level of quality of experience, and as over time the target level changes, the agent should seamlessly continue to pursue the new goal, i.e., the agent generalizes over the domain of goals. One way of achieving such results is via goal-conditioned reinforcement learning.

%In goal-oriented environments, such as the one described above, the goal states are known beforehand. Concretely, incoming intents already specify the target KPI level, so that the agents immediately know if their measured KPI value is above or below such target. This characteristics also allow us to  specify reward functions based on the similarity between the observed and the target value.

These aspects can be formalized by state-action value functions $Q(o, g, a)$, where $o$ and $g$ come from the KPI domain and reward functions $r(o, g) = d(o, g)$, where $d$ refers to any similarity measure~\cite{schaul2015uvf}. During training, $g$ values are randomly chosen from the domain at the beginning of every episode, so that the resulting policy is able to generalize over different goals during the execution phase. 

The simpler way of implementing goal-conditioned \ac{RL} is by adding the goals $g$ as an additional dimension in the observation space. That effectively changes the observation space, which we formalize later. Although there might be limitations to such an approach~\cite{schaul2015uvf}, it allows us to easily employ out-of-the-box \ac{RL} algorithms in a goal-conditioned setting.

\section{Methodology}
\subsection{Network Emulator}
For the training environment, a network emulator was employed with the basic requirements
for the problem at hand. The emulator provides an end-to-end environment, from compute resources
hosting services (e.g., edge and central sites) to the users of those services. The emulator
models gNBs and \acp{UPF} connecting the \acp{UE} to the
services. Figure 1 illustrates the network topology implemented on the emulator.

Three service types are supported on the current version of the emulator - i.e. 
\ac{CV}, \ac{URLLC} and \ac{mIoT} services. 
Services can serve multiple \acp{UE} at the same time, while \ac{URLLC} and \ac{mIoT} services
can have multiple instances. The emulator exposes a collection of \ac{API} endpoints. 
The APIs give the agents access to several data points, including Throughput \ac{UL}/\ac{DL},
Packet Loss \ac{UL}/\ac{DL}, Number of Bytes \ac{UL}/\ac{DL}, services' packet priority, \acp{UE} \ac{MBR}, 
Service latency and Service \ac{QoE}. It also allow the agents to act on the environment by modifying
individual \acp{UE} \ac{MBR} and Services' packet priority on the network. To handle interactions 
with the network emulator a gym-like interface was implemented and exposed to the agents.

%\subsubsection{Gym Interface}: To handle interaction with the environment, a gym-like 
%environment was developed encapsulating the network emulator and exposing relevant data and
%actions through well-known interfaces. A multi-agent environment is defined where multiple actions
%can be taken simultaneously which in turn yields agent specific observations and reward calculations.
%The interface provides a \emph{step} function that takes agents' joint actions as input and returns
%the joint observations and rewards for each agent.

\begin{figure*}[!t]
    \centering
    \includegraphics[width=0.9\textwidth]{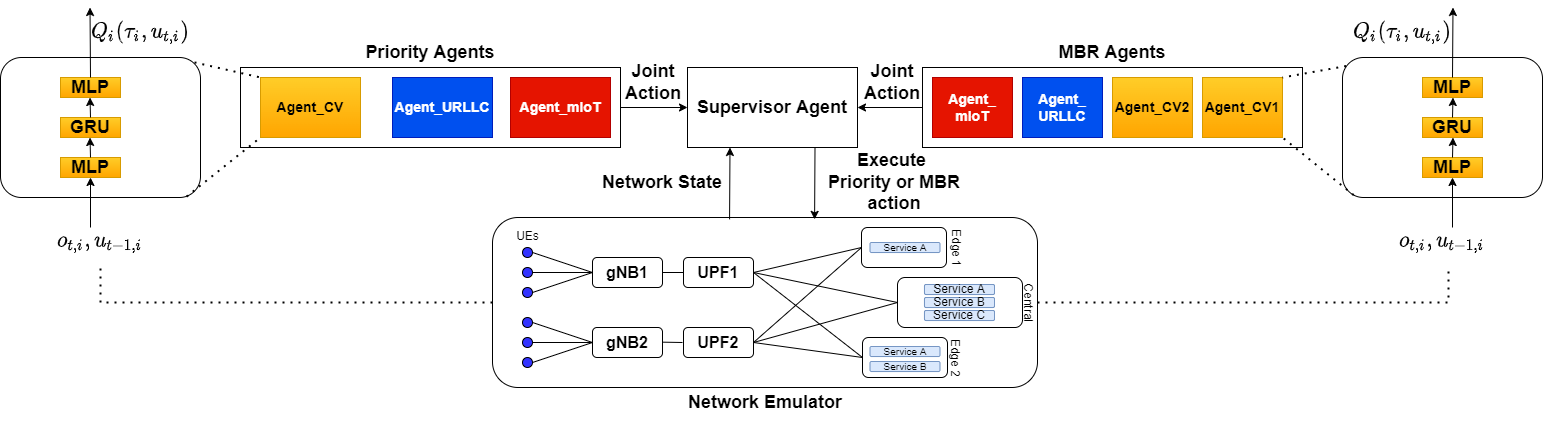}
    \caption{MARL setup and environment.}
    \label{fig:MARL}
\end{figure*}
%\subsection{Env - Network emulator, MBR , priority}

\subsection{MARL Environment specification}

In our scenarios, multiple closed loops are interacting while optimizing their particular \acp{KPI} by tuning particular control knobs. Those characteristics translate to \ac{MARL} formalism as multiple heterogeneous agents for which the conflicting demands for resources require cooperative behavior. 

In this work, we employ traditional QMiX~\cite{rashid2018qmix} to train agents able to cooperate without communication in decentralized settings. The objective is to arrive at goal-conditioned agents able to adapt to dynamic changes in the intents while also maintaining high performance in an open MARL setup, where agents may join or leave at any time. Next, we specify the \ac{RL} components defined on top of the network emulator infrastructure.

 \subsubsection{Observation space}
Since we deal with multiple heterogeneous agents, we first specify the local observation spaces, and later the joint or global observation space. 

Each service type is associated with a single \ac{KPI}, e.g., while the quality of a conversational video service is measured in terms of \ac{QoE}, the quality of an \ac{URLLC} service might be measured in terms of packet losses. Overall, the first dimension of the local observation spaces is always the respective \ac{KPI} measurement, while the second dimension is the \ac{KPI} target defined by the intents. To facilitate the emergence of cooperation during training, we also include the global reward $G\in\mathbb{R}$ and the number of \acp{UE} $n\in\mathbb{N}$ using the service as additional dimensions. 

Considering $T$ as a service type index, and $\mathcal{K}_T$ as the \ac{KPI} ranges associated with $T$, the local observation spaces are defined as $\mathcal{S}_T \subseteq \mathcal{K}_T\times \mathcal{K}_T\times \mathbb{R}\times \mathbb{N}$. The specific \ac{KPI} and their domains are specified in Table~\ref{tab:obs}. For easy notation, we consider $s=(o_j, g_j, G, n_j)\in\mathcal{S}_j$, where $j$ is an index referring to the service types, and $o_j, g_j\in\mathcal{K}_j$.
\begin{table}[h]
    \centering
    \caption{Service types, KPIs and domains.}
    \begin{tabular}{llll}
        \toprule
        \# & Service type $T$ & KPI & $\mathcal{K}_T$\\\midrule
        1 & Conversational video (CV) & \ac{QoE} & $[1,5]$\\
        2 & URLLC & Packet-loss ratio (PLR) & $[0,1]$\\
        3 & mIoT & Packet-loss ratio (PLR) & $[0,1]$\\\bottomrule
    \end{tabular}%
    \label{tab:obs}
\end{table}

%\begin{enumerate}
%    \item Conversational Video Service: $(QoE, QoE-Target_{QoE},Global_reward,n_{cv})$
%    \item URLLC Service: $(PL_{URLLC},PL_{URLLC}-Target_{PLURLLC},Global_reward,n_{urllc})$
%    \item mIoT Service: $(PL_{mIoT},PL_{mIoT}-Target_{PLmIoT},n_{mIoT})$
%\end{enumerate}
%where $n_{.}$ is the number of UE's in the service $.$. In the above set of observations the $Global_reward$ is added to all the agent's local observations to ensure good collaboration as each agent has to reach their own targets with respecting to the overall target reaching capability of all the agents. 

Altogether, the local observation spaces compose a joint observation space that can be interpreted as the global state of the \ac{MARL} environment. Such joint observation space is based on the concatenation of all local observation spaces, which is composed of elements $\mathbf{s}\in\boldsymbol{\mathcal{S}}$.

%\textbf{State}: Since we have heterogeneous agents, we combine the current value of the parameters which the agents are controlling i.e. at time instant $k$ state of the environment is measured as 
%\begin{align}
%    State[k] = (QoE[k],PL_{URLLC}[k],PL_{mIoT}[k])
%\end{align}

\subsubsection{Reward function}
Analogous to the observation spaces, the reward functions are also divided in local and global. The local reward functions concerns the individual \acp{KPI} domains, and are defined to comply with the goal-conditioning setup, i.e., $r_j(o,g)=|o - g|$, where $o,g\in\mathcal{K}_T$. Such reward functions would output higher values as the current \ac{KPI} measurement $s$ is further away from the target $g$ in either direction. 

Additionally, as we are dealing with heterogeneous agents and each reward function may be in a different range, we scale down the individual rewards into the range $[0,1]$ by employing a normalization factor $\Delta_j$ that represents the maximum absolute difference between $o_j$ and $g_j$. Therefore, given a pair $(o_j, g_j)$, the reward function for a service $j\in T$ is define as
\begin{align}
    r_j(o_j, g_j) = 1 - (|o_j - g_j|/\Delta_j).
\end{align}

From local rewards functions, local observations and goals, a  linear global reward function can be defined. Here, we also introduce the notion of penalties $\rho_j$ (or preferences). If all penalties are equal, all services should be equally served, otherwise, those services with higher penalties should be favored at the expense of the others.
\begin{align}
    G(\mathbf{s}) = \sum_{s_j\in \mathbf{s}} \rho_j \cdot r_j(o_j, g_j)
\end{align}

%\begin{align}
%    Global_{reward} = & ((QoE-Target_{QoE})/Target_{QoE})+ \nonumber\\ & %((PL_{URLLC}-Target_{PLURLLC})/Target_{PLURLLC})+ \nonumber\\ & %((PL_{mIoT}-Target_{PLmIoT})/Target_{PLmIoT})
%\end{align}

Penalties are particularly important whenever we don't have enough resources to achieve all intents. In such cases, equal penalties lead to equal degradation of the service KPIs. By adding the penalty terms to the global reward function, we can tune the desired behavior by properly choosing different penalty terms for each service.

%In some cases where we don't have enough resources present to achieve the intents, the MARL can result in equal degradation owing to reward function chosen above. However, in some case we have penalty present in intents which suggests a specific KPI to be prioritized. Hence, we modify the reward function to include the penalties to train the agents when there are scarce resources i.e. individual KPI's are multiplied by respective penalty values.  

All the above components are same for both the packet priority MARL agent and MBR MARL agent. However, their action space is different and is explained below.  

\subsubsection{Action spaces}
Service assurance for each service type can be pursued via tuning of multiple configuration parameters (control knobs), i.e., for each service we can have multiple agents. In this paper, we consider two control knobs the \ac{MBR} and the packet priority. Therefore, we have two types of agents which we refer to as \ac{MBR} agents and packet priority agents, both learned via \ac{RL}. Since QMIX was our algorithm of choice, both action spaces were enforced to be discrete. 

In the case of \ac{MBR} agents, the actions consist of decisions for increasing or decreasing the current \ac{MBR} value, i.e., $\mathcal{A}=\{-1,1\}$. The increment/decrements are fixed and defined from the discretization of the original action space $[1, \alpha]$ into bins of size $0.5$ Mbps (where $\alpha$ is the air-link bandwidth). In this setting an increase/decrease action  would move to the next/previous bin. After the updated bin $b_t =  b_{t-1} + a_t$ is identified, a random value is chosen within its bounds to specify the \ac{UE}'s \ac{MBR}.
\begin{align}
    \mbox{MBR}_t &= \mbox{rand}(b_t)
\end{align}

%For the MBR loop we divided the entire range of MBR i.e. from $0$ to Air link bandwidth into different bins of length $0.5 \; Mbps$ each. Now, the agent has to decide whether to move to the left bin or right bin from the existing bin. In this way, we can increase or decrease MBR for the agent. It should be noted that in any of the bin, we select a random value from the bin and use it as MBR for corresponding UE. 

Similarly, in the case of priority agents, the actions consist of decisions for increasing or decreasing the current packet priority of a service, i.e., $\mathcal{A}=\{-1,1\}$. The minimum/maximum priority values are defined to comply with the Network Emulator design, e.g., $[1, 100]$, where a small value denotes high priority. 
\begin{align}
    \mbox{Priority}_t = \mbox{Priority}_{t-1} + a_t
\end{align}

%For the priority loop we consider two actions i.e. to increase priority or decrease priority. To minimize the action space here, we consider the priority range between $1$ to $100$.

%The action space used in this work is 
%\begin{enumerate}
%    \item MBR Action: $MBR[k] = MBR[k-1] + Action[k] *(0,0.5]$
%    \item Priority Action: $priority[k] = priority[k-1]+Action[k]$
%\end{enumerate}

\subsection{MARL training and evaluation}

Each agent is modelled by 2-layer \ac{GRU} network with $2$ nodes in each layer. During training, QMIX benefits from local and global observations and rewards, i.e., centralized training. The training phase consists of multiple episodes, each lasting for a maximum of $H=30$ time steps or until all agents have reached their intents. During the evaluation phase, agents only rely on their local observations to reach their individual goals, i.e., decentralized execution. 

Packet priority and \ac{MBR} agents have different scopes, as specified by the network emulator used for the experiments. While packet priority is defined at the service level, \ac{MBR} is defined at the \ac{UE} level. Although it is scalable to set packet priority for each service, it may not be realistic to set \ac{MBR} to individual \acp{UE}. Therefore, \ac{MBR} agents take decisions for sets of \acp{UE} instead. 

%The network emulator is designed with a functionality to set MBR values at the UE level. Hence, we decided to operate the MBR action at the UE level. However, it should be noted that considering each UE as one agent increases overall complexity and hence we opted to consider group of the UE's as one agent. The idea behind this is explained below. 

The rationale for grouping UEs depend on the format of the intents supported. Each intent follows a scheme \textit{'$x\%$ of UE's to have KPI greater than $g$'}. Hence, we can divide the UEs from service into two groups of sizes $x\%$ and $(100-x)\%$. The \ac{MBR} agents' actions then affect all the UEs within a group equally, all of them getting the same MBR value. If the intent is on 100\% of UEs then the MBR actuation reduces to the service level,  analogously to packet priority agents.

Both priority and MBR agent groups were trained independently to reduce the computational overhead. Additionally, we assume that they never actuate at the same time. Given that the agents have different strengths and weaknesses, such an approach requires an additional level of decision-making specifying which agents to activate and deactivate during execution time. In this paper, we evaluate a very simple supervisor agent, which chooses the specific agent to activate based on knowledge about how the network emulator works
\begin{table}[htb]
\caption{Supervisor agent and its rules.}
    \centering
    \begin{tabular}{lll}
        \toprule
         \# & Rule & Agent to use \\\midrule
         A. & If all UEs throughput $=$ MBR, & MBR agents \\
         B. & Otherwise, & Priority agents\\
         \bottomrule
    \end{tabular}
    \label{tab:rules}
\end{table}

The supervisor agent decisions take place every 5 time steps, after which the chosen agent (MBR or Priority) stay active for another round. The overall architecture of the proposed method is illustrated by Figure \ref{fig:MARL}. 
%\begin{figure}[htb]
%    \centering
%    \includegraphics[width=\columnwidth]{MARL.png}
%    \caption{MARL methodology used in this work. }
%    \label{fig:MARL}
%\end{figure}
%The supervisor agent runs for every 5 seconds and decides which agent to deploy and use the specific agent for next 5 seconds. In this way, we obtain an efficient usage of both the agents. 

%In the next section, we discuss the results of applying proposed method on the emulator in reaching the intents supplied by the customers. 

\section{Results and Discussions}

We evaluated the efficacy of the proposed solution in an end-to-end network emulator, considering two scenarios, (1) the first considers that plenty of resources are available, allowing the fulfilment of all intents, while (2) in the second resources are scarce. To assess the benefits and limitations of the supervisor agent we also compare the results with those achieved by employing only priority or only MBR agents. 
\subsection{Experimental setting}

\subsubsection{Performance metric}
To compare and quantify the methods w.r.t goal realization and convergence speed we use a metric defined in \eqref{eq:M}. The metric computes the average distance between observed KPI values to the intended target. 
%To quantify how close to the incoming intents the different approaches could go, and also how fast convergence towards the goals is, we specify a metric which measures the average distance of observed KPI values to the intended KPI targets, which is defined by Equation~\eqref{eq:M}. 
\begin{align}
    M = \frac{1}{H}\displaystyle \sum_{t=1}^{H} |o_t-g_t|
    \label{eq:M}
\end{align}

where $H$ is number of time steps, $o_t$ is the observed KPI value and $g_t$ is the intended KPI value (goal) at instant $t$. An ideal agent would produce trajectories whose $M$ is close to zero, i.e., the agent  quickly reaches its goals and stay close to it during its whole execution time. 

\subsubsection{Intents specification}
We use the same set of intents for both scenarios (plenty vs scarce resources). After a period of time, we induce a change in CV service intent (from 1.1 to 1.2 in Table below) to evaluate agent's ability to adapt. This is where goal-conditioned training becomes beneficial. The whole set of intents is defined below.
% \begin{itemize}
% \item $75\%$ of UEs from CV should have QoE $\geq 3$.
% \item $75\%$ of UEs from CV should have QoE $\mathbf{\geq 3.5}$. 
% \item $100\%$ of UEs from URLLC should have PLR\footnote{Packet-loss ratio.} $\leq 2\%$.
% \item $100\%$ of UEs from mIoT should have PLR $\leq 2\%$.
% \end{itemize}
\begin{table}[htb]
    \centering
    \begin{tabular}{lcccc}
        \toprule
         Intent & Service & \% of UEs & KPI & KPI target\\         \midrule
         1.1 & CV    &  75 & QoE & $\geq 3.0$ \\
         1.2 & CV    &  75 & QoE& $\geq 3.5$ \\
         2 & URLLC & 100 & Packet-loss ratio & $\leq 0.02$ \\
         3 & mIoT  & 100 & Packet-loss ratio & $\leq 0.04$\\
         \bottomrule 
    \end{tabular}
    \label{tab:intents}
\end{table}

\subsubsection{Network emulator settings}
We configured 4 UEs for each service and 2 gNBs connected to an equal number of UEs across all the services. At the beginning of an episode, all services are set to the same packet priority ($7$), while all UEs are set to the same MBR value (1Mbps). 

Once the agents change the priority or MBR values, the effect is not immediately observed. Instead, the effect of actions is delayed by a certain period of time that depends of the type of action and the network state. In this paper, we defined a fixed time window of 40 seconds before observing the effect of packet priority changes, while a time window of 10 seconds for MBR changes. Those values were estimated in preliminary experiments which we omit here for brevity. 

\subsection{Results}
In this section we illustrate results for three agent group setups: (a) Only priority agents, (b) only MBR agents, and (c) both agents orchestrated via a Supervisor. For brevity, we summarize the overall performance results in tables and show plots only for the third setup.

\subsubsection{Scenario 1: Plenty of Resources}
In this scenario, the airlink bandwidth is 20 Mbps, which is enough for all intents being met (there is no bottleneck in the transport layer).

%The agents in priority loop are converged after 100 episodes with average episode length of 20 time steps, whereas agents in MBR loop are converged after 68 episodes with average episode length of 18 time steps.  As mentioned in previous paragraphs, each time step corresponds to 40 seconds in priority loop and 10 seconds in MBR loop. 

\begin{figure}[htb]
    \centering
    \includegraphics[width=\columnwidth]{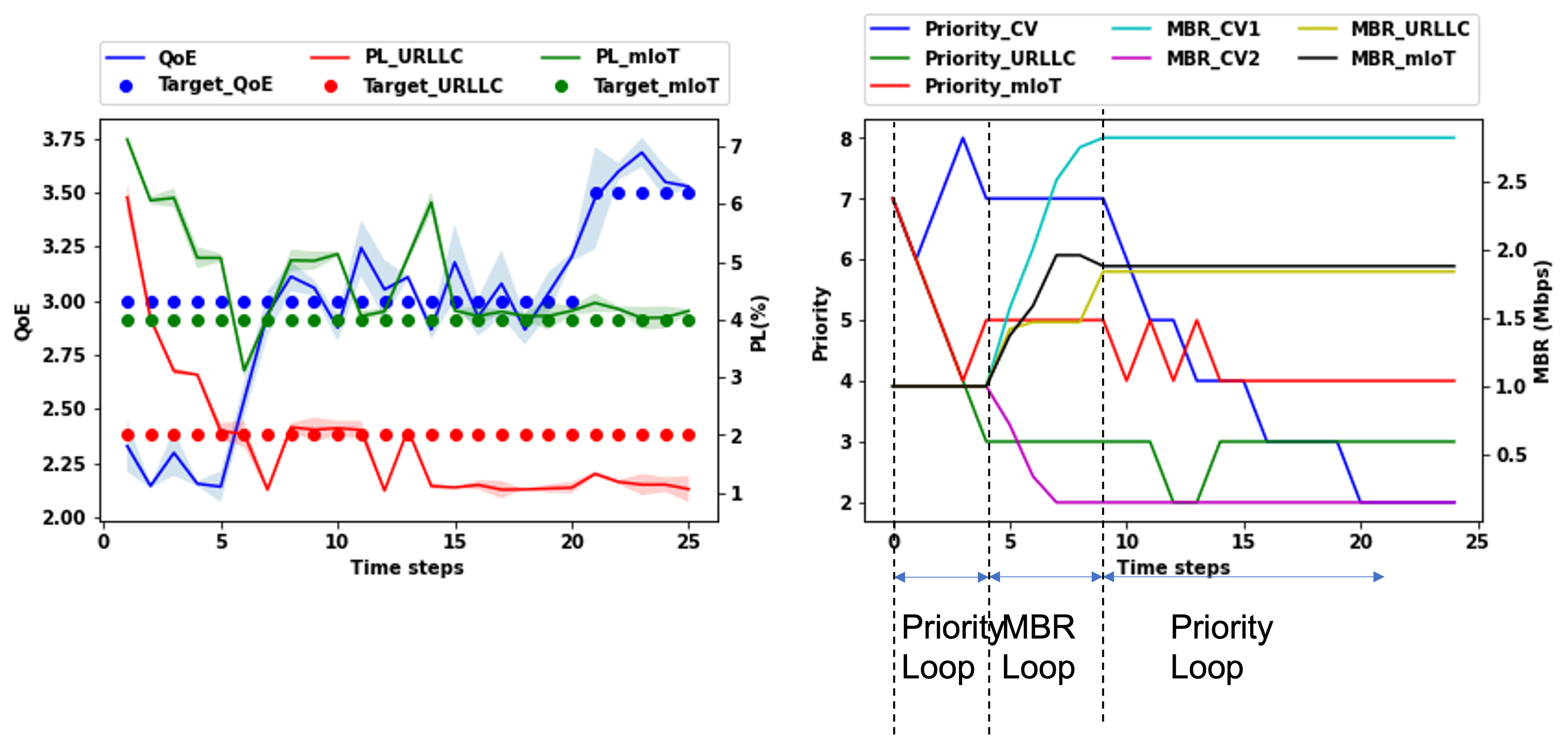}
    \caption{Results with a supervisor agent and plenty of resources. }
    \label{fig:sup_available}
\end{figure}

Figure \ref{fig:sup_available} shows the results for this scenario. The figure at left shows how QoE and Packet-loss ratio (lines) reach the KPI targets (dots) defined by the intents. The figure at the right shows how the agents' actions affect priority and MBR values over time. Here, we can also observe the time windows in which each agent is active (e.g., packet priority loop, MBR loop). From the plots, it is evident that how the agents are able to readjust their trajectory even if the goals/intent changes midway. Hence given the agents are goal-conditioned they can reach any goal value from any point in state space. 

The performance results $M$ for all agent group setups are summarized in Table \ref{tab:available}. The smaller values indicate that the orchestration of priority and MBR MARL loops, with the help of a supervisor agent, led to faster and better convergence towards the intended goals. Here, we note that the bad performance of MBR agents for mIoT services is expected, since such services require very low throughput and do not benefit much from MBR increases. 

\begin{table}[htb]
    \centering
    \caption{$M$ value computed for available resources}
    %\begin{tabular}{|c|c|c|c|}
    \begin{tabular}{lccc}
        \toprule
         \textbf{Service} & \textbf{Only Priority} &\textbf{Only MBR} & \textbf{Priority \& MBR}\\         \midrule
         CV &  1.49 & 0.97 & 0.25 \\
         URLLC & 1.27 & 1.68 & 0.85 \\
         mIoT & 1.49& 2.27 & 1.09\\
         \bottomrule 
    \end{tabular}
    \label{tab:available}
\end{table}
\subsection{Scenario 2: Scarce Resources}
In this scenario, the airlink bandwidth is 4 Mbps, which is not enough for all intents to be met.

%In this case, the agents in priority loop are converged after 129 episodes with average episode length of 28 time steps. On the other hand, agents in MBR loops converges after 85 episodes with average episode length of 30 time steps. 

\begin{figure}[htb]
    \centering
    \includegraphics[width=\columnwidth]{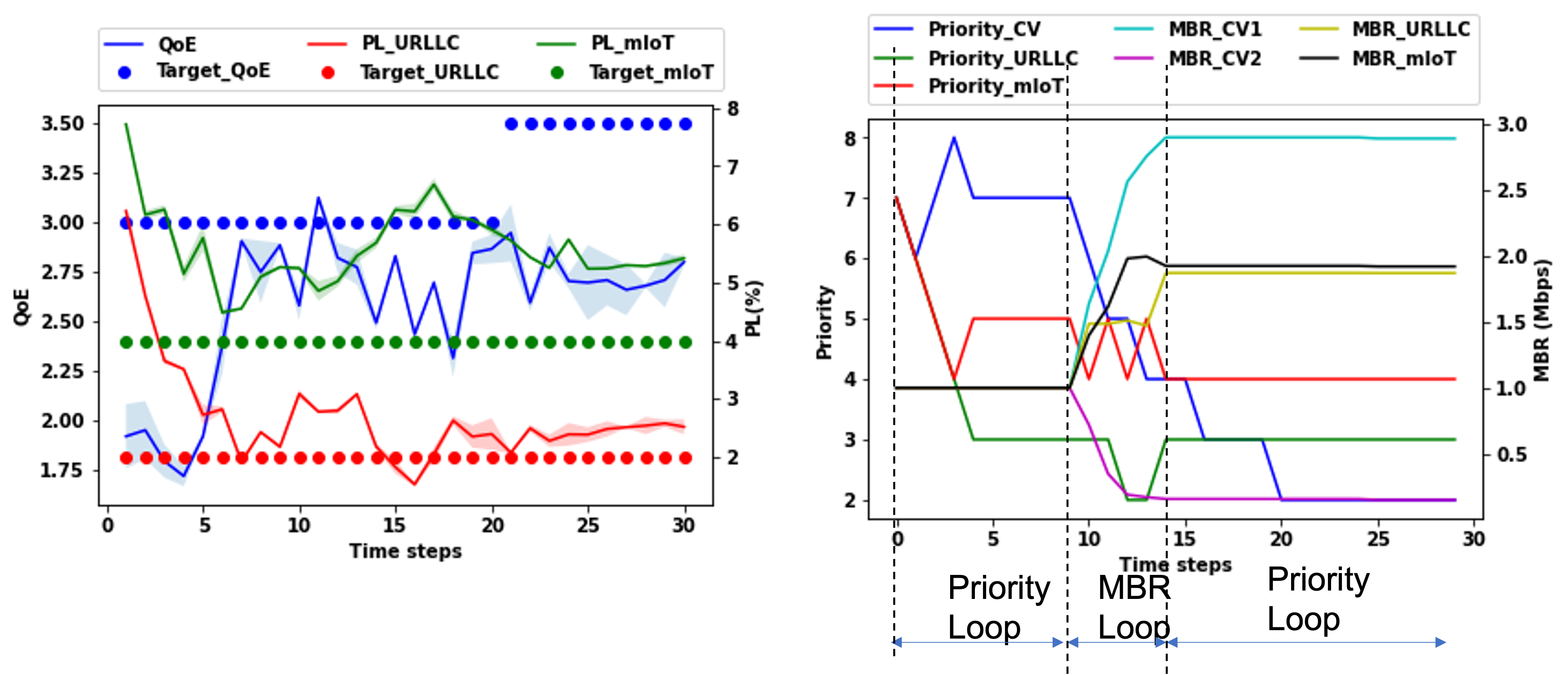}
    \caption{Results with a supervisor agent, scarce resources and equal penalties. }
    \label{fig:sup_scarce}
\end{figure}

Figure \ref{fig:sup_scarce} show the results for this scenario. The supervisor agent decides to activate the priority agents twice in first ten time steps, followed by the MBR agents for next five time steps, and then again the priority agents. Here we notice that, due to the scarcity of resources, all observed KPIs are below/above their targets and intents are not met. The performance results $M$ are summarized in Table \ref{tab:scarce}. 

\begin{table}[htb]
    \centering
    \caption{$M$ value computed for scarce resources}
    \begin{tabular}{lccc}
    \toprule
         \textbf{Service} & \textbf{Only Priority} &\textbf{Only MBR} & \textbf{Priority \& MBR} \\\midrule
         CV &  1.58 & 1.24 & 0.64 \\
         URLLC & 1.07 & 1.74 & 0.94 \\
         mIoT & 1.84 & 2.45 & 1.94\\
         \bottomrule
    \end{tabular}
    \label{tab:scarce}
\end{table}

However, the behavior of equally degrading all services performance may not be desired in all scenarios. From the business perspective, sometimes a specific service has to be prioritized over others even when there are not enough resources. Let us assume, URLLC service, for example, is assigned as high priority i.e. it has to reach its intent even at cost of degrading the others. To achieve such autonomous trade-offs for specific services in scarce resources scenarios, we introduce differentiated penalties for each service, to quantify how the violation of a high priority service intent compares against others. Therefore, to prioritize URLLC over CV and mIoT we associate a higher penalty to it, e.g., $\rho=10$ for URLLC and $\rho=1$ to CV. By considering such penalties in the global component of the reward functions, the agents learn to prioritize the URLLC service over others. 

\begin{figure}[htb]
    \centering
    \includegraphics[width=\columnwidth]{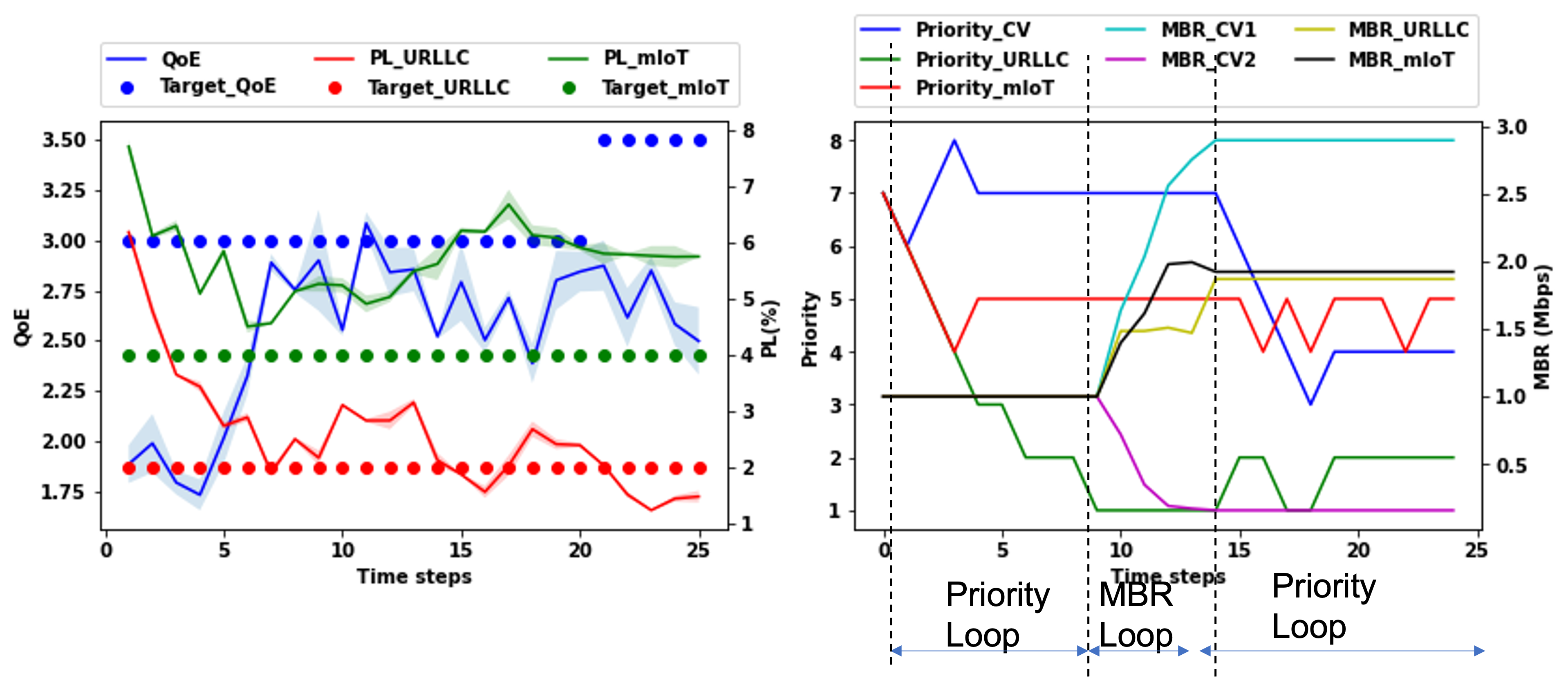}
    \caption{Results with a supervisor agent, scarce resources and differentiated penalties. Agents learn to trade-off for a prioritized service.  }
    \label{fig:penalty}
\end{figure}

Figure \ref{fig:penalty} show the results for scarce resources with differentiated penalties. From these results we observe that the URLLC Packet-loss ratio now achieves its intent, which indicates that penalties are a valid approach for balancing service preferences. Also, from the plot it can be observed that the other two services got degraded, i.e learn to trade-off, to enable URLLC service reach its intent. 

%From the results it can be seen that URLLC service reached its intent faster owing to higher penalty than other services. In this way, any service can be prioritized by giving them a higher penalty value to it. 

%However, the proposed approach of training the agents works better for a interval of penalty values without any additional re-training. Currently, we are working to make agents generic to wide range of penalty values and will be part of future work. 

%*CAN BE DELETEDFrom these two cases, it can be seen that proposed approach can be used to reach any intents from any values. When there are scarce resources, all KPIs will degrade equally as they have equal penalty. With the help of penalty we can make a specific KPI reach its intent faster than other KPIs. 

\section{Conclusions} 
In this paper, we proposed a method based on MARL to achieve multiple intents (KPI's) in a interacting closed loops scenario. Additionally, we formulated goal-conditioned MARL agents which helps to generalize agents' ability to seek goals across a wide range of values. The proposed approach is tested on a network emulator for three services CV, URLLC and mIoT for varying amount of resource availability. The experiments demonstrate that agents can learn the "Plan to Coordinate" in order to manage conflicts and promote cooperation towards maximizing the global goal. When there are enough resources available, the trained MARL agents coordinate to achieve all intents. Also, when resources are scarce, based on intent priorities (denoted by penalties in this work), the agents learn an optimal trade-off mechanism, which resulted in proportional degradation of an intent to prioritize another intent of higher priority. Finally with use of supervisor agent, we demonstrate a method to autonomously coordinate multiple MARL loops which could later be extended to hierarchical control across domains.

Future directions include extending the approach to larger number of agents, designing generic agents and testing the approaches with different reward functions.
\bibliographystyle{IEEEtran}
\bibliography{references}

\begin{thebibliography}{10}
\providecommand{\url}[1]{#1}
\csname url@samestyle\endcsname
\providecommand{\newblock}{\relax}
\providecommand{\bibinfo}[2]{#2}
\providecommand{\BIBentrySTDinterwordspacing}{\spaceskip=0pt\relax}
\providecommand{\BIBentryALTinterwordstretchfactor}{4}
\providecommand{\BIBentryALTinterwordspacing}{\spaceskip=\fontdimen2\font plus
\BIBentryALTinterwordstretchfactor\fontdimen3\font minus
  \fontdimen4\font\relax}
\providecommand{\BIBforeignlanguage}[2]{{%
\expandafter\ifx\csname l@#1\endcsname\relax
\typeout{** WARNING: IEEEtran.bst: No hyphenation pattern has been}%
\typeout{** loaded for the language `#1'. Using the pattern for}%
\typeout{** the default language instead.}%
\else
\language=\csname l@#1\endcsname
\fi
#2}}
\providecommand{\BIBdecl}{\relax}
\BIBdecl

\bibitem{paper:moysen_conflict_res}
J.~Moysen, M.~Garcia-Lozano, L.~Giupponi, and S.~Ruiz, ``Conflict resolution in
  mobile networks: A self-coordination framework based on non-dominated
  solutions and machine learning for data analytics,'' \emph{IEEE Computational
  Intelligence Magazine}, vol.~13, no.~2, pp. 52--64, May 2018.

\bibitem{report:Ericsson}
\BIBentryALTinterwordspacing
J.~Niemoller, L.~Mokrushin, S.~Mohalik, V.~Konchylaki, and Sarmonikas,
  ``Cognitive processes for adaptive intent-based networking,'' Ericsson
  Technology Review, Kista, Sweden, Tech. Rep., 2020. [Online]. Available:
  \url{https://www.ericsson.com/4abd97/assets/local/reports-papers/ericsson-technology-review/docs/2020/adaptive-intent-based-networking.pdf}
\BIBentrySTDinterwordspacing

\bibitem{tmforumIntents}
{TM Forum Autonomous Network Project}, ``Ig1253 intent autonomous networks,''
  https://www.tmforum.org/resources/how-to-guide/ig1253-intent-in-autonomous-networks-v1-1-0/.

\bibitem{etsi-zsm-closedloops}
{ETSI GS ZSM 009-1}, ``Zero-touch network and service management (zsm):
  Closed-loop automation; enablers group specification.''

\bibitem{moysen2018}
J.~Moysen and L.~Giupponi, ``From 4g to 5g: Self-organized network management
  meets machine learning,'' \emph{Computer Communications}, vol. 129, pp.
  248--268, 2018.

\bibitem{munoz2014}
P.~Mu$\tilde{\hbox{n}}$oz, R.~Barco, and S.~Fortes, ``Conflict resolution
  between load balancing and handover optimization in lte networks,''
  \emph{IEEE Communications Letters}, vol.~18, no.~10, pp. 1795--1798, 2014.

\bibitem{rojas2020}
D.~F. Preciado~Rojas and A.~Mitschele-Thiel, ``A scalable son coordination
  framework for 5g,'' in \emph{NOMS 2020 - 2020 IEEE/IFIP Network Operations
  and Management Symposium}, 2020, pp. 1--8.

\bibitem{iacoboaiea2014}
O.~Iacoboaiea, B.~Sayrac, S.~Ben~Jemaa, and P.~Bianchi, ``Son coordination for
  parameter conflict resolution: A reinforcement learning framework,'' in
  \emph{2014 IEEE Wireless Communications and Networking Conference Workshops
  (WCNCW)}, 2014, pp. 196--201.

\bibitem{sutton1998introduction}
R.~S. Sutton and A.~G. Barto, \emph{Introduction to reinforcement
  learning}.\hskip 1em plus 0.5em minus 0.4em\relax MIT press Cambridge, 1998,
  vol. 135.

\bibitem{paper:iql}
M.~Tan, ``Multi-agent reinforcement learning: Independent vs. cooperative
  agents,'' in \emph{Proceedings of the tenth international conference on
  machine learning}, 1993, pp. 330--337.

\bibitem{rashid2018qmix}
T.~Rashid, M.~Samvelyan, C.~Schroeder, G.~Farquhar, J.~Foerster, and
  S.~Whiteson, ``{QMIX}: Monotonic value function factorisation for deep
  multi-agent reinforcement learning,'' in \emph{Proceedings of the 35th
  International Conference on Machine Learning}, vol.~80, 10--15 Jul 2018, pp.
  4295--4304.

\bibitem{rashid2020qmix}
T.~Rashid, M.~Samvelyan, C.~S. de~Witt, G.~Farquhar, J.~Foerster, and
  S.~Whiteson, ``Monotonic value function factorisation for deep multi-agent
  reinforcement learning,'' \emph{Journal of Machine Learning Research},
  vol.~21, no. 178, pp. 1--51, 2020.

\bibitem{schaul2015uvf}
T.~Schaul, D.~Horgan, K.~Gregor, and D.~Silver, ``Universal value function
  approximators,'' in \emph{Proceedings of the 32nd International Conference on
  Machine Learning}, ser. Proceedings of Machine Learning Research, F.~Bach and
  D.~Blei, Eds., vol.~37.\hskip 1em plus 0.5em minus 0.4em\relax Lille, France:
  PMLR, 07--09 Jul 2015, pp. 1312--1320.

\end{thebibliography}

\end{document}